\documentclass[10pt]{article}
\usepackage[a4paper, textwidth=16cm]{geometry}
\usepackage{multirow,multicol}
\usepackage{amsmath}
\DeclareMathOperator*{\argmax}{arg\,max}
\usepackage{amsthm}
\usepackage{mathtools}
\usepackage{amssymb}
\usepackage{caption}
\usepackage{bm}

\usepackage{titlesec}
\newtheorem{theorem}{Theorem}[section]
\newtheorem{definition}[theorem]{Definition}
\newtheorem{example}[theorem]{Example}
\DeclareMathOperator{\diag}{diag}

\usepackage{mhchem,multirow,calc,array,amsmath,booktabs}
\newlength\mylenA
\newlength\mylenB
\newlength\mylenC

\usepackage{color}

\titleformat{\chapter}[display]
  {\normalfont\bfseries}{}{0pt}{\Huge}

\title{A classification model based on\\
a population of hypergraphs}
\usepackage{authblk}

\newsavebox\affbox
\author[1]{\textbf{Samuel Barton}}
\author[2]{\textbf{Adelle Coster}}
\author[1]{\textbf{Diane Donovan}}
\author[1]{\textbf{James Lefevre}}
\affil[1]{ARC Centre of Excellence, Plant Success in Nature and Agriculture, School of Mathematics and Physics,  University of Queensland, Brisbane, 4072, Australia}
\affil[2]{ School of Mathematics \& Statistics, The University of New South Wales, NSW 2052, Australia}
\begin{document}

\maketitle
\section*{Abstract}
This paper introduces a novel hypergraph classification algorithm. The use of hypergraphs in this framework has been widely studied. In previous work, hypergraph models are typically constructed using distance or attribute based methods. That is, hyperedges are generated by connecting a set of samples which are within a certain distance or have a common attribute. These methods however, do not often focus on multi-way interactions directly. The algorithm provided in this paper looks to address this problem by constructing hypergraphs which explore multi-way interactions of any order. We also increase the performance and robustness of the algorithm by using a population of hypergraphs. The algorithm is evaluated on two datasets, demonstrating promising performance compared to a generic random forest classification algorithm. 

\newpage
\section{Introduction}
In recent years, the field of mathematics and computer science has witnessed an increased interest in hypergraphs and their diverse applications. Hypergraphs, as a generalisation of graphs, offer a versatile framework for modeling complex relationships. This has led to growing significance in various domains, including social networks \cite{li2013link}, bioinformatics \cite{di2021hypergraph}, and pattern recognition \cite{zhou2006learning}. An interesting overlap between hypergraph theory and computer science is their application in classification algorithms.

Hypergraphs extend the concept of graphs by allowing edges to connect not just pairs of vertices, but sets of vertices, which are often referred to as hyperedges. This structure enables hypergraphs to capture more intricate and higher-order relationships, making them a useful tool for representing data with complex and multi-way interactions.

Classification algorithms, referred to as classifiers, are a core concept in machine learning, enabling the categorisation of data units into distinct classes based on features of the data. Specifically, these types of algorithms use data with class labels to train a model that will predict the mostly likely class for unlabeled data points. Leveraging hypergraphs in classification algorithms can yield several advantages. Hypergraphs can encode higher-order relationships that might be crucial for accurate classification. They can capture mutually exclusive relationships among features, which often go unnoticed when using conventional graphs. Moreover, hypergraphs provide a more flexible way to handle data with irregular structures, where relationships among multiple entities are vital for informed decision-making. This is exemplified in a range of approaches \cite{bai2021hypergraph,feng2019hypergraph,sun2008hypergraph,wang2015visual,yu2012adaptive}.  

The aim of this research paper is to formulate a novel hypergraph based classifier and to study the performance on a starch grain data set. We also present the hypergraph classifier in such a way that it can be applied to a range of datasets. By harnessing the inherent structure of hypergraphs, these algorithms have the potential to improve classification accuracy and feature selection. Specifically, the hypergraph framework enables the inclusion of higher order feature interactions which is shown to increase predictive performance. A rise in predictive performance is also achieved through a population of hypergraph models. In turn, this creates a flexible algorithm which can be generalised easily to any data set. The introduction of a threshold also increases performance, where only half of the data units are classified but with an increased accuracy. 

The subsequent sections of this paper are organized as follows: Section 2 provides the necessary background on hypergraph theory and linear algebra, highlighting key definitions and properties. In Section 3, we derive the hypergraph based classification algorithms. Section 4 explores case studies, applying the novel hypergraph classifier and comparing the performance to benchmark classifiers. In this section we also discuss various threshold techniques which can be used to increase performance. Finally, Section 5 concludes the paper by summarising the contributions and offering insights into the  future directions of hypergraph-based classification algorithms.

\section{Mathematical Preliminaries}
We first briefly introduce the relevant definitions of hypergraphs, a generalisation of graphs. A hypergraph $\mathcal{H = (V,E)}$, refers to a vertex set $\mathcal{V} = \{v_1,v_2,\dots,v_n\}$ and hyperedge set $\mathcal{E} = \{e_1,e_2,\dots,e_m\}$, where each hyperedge corresponds to a subset of vertices of $\mathcal{V}$ \cite{berge1984}. 

The number of vertices is the \textit{order} of the hypergraph and number of hyperedges is the \textit{size}. We only consider unweighted and undirected hypergraphs, where order is greater than zero. The \textit{cardinality} or \textit{size} of a particular hyperedge is the number of vertices in that hyperedge. We say two vertices are \textit{adjacent} if they are elements of the same hyperedge. A hyperedge and vertex are said to be \textit{incident} if the vertex is an element of the hyperedge. The \textit{intersection} of two or more hyperedges is the set of common vertices. 

An \textit{incidence matrix} provides a concise and unique representation for a hypergraph. Formally, the incidence matrix $B = [b_{i,j}]$ of hypergraph $\mathcal{H = (V,E)}$, $\mathcal{V} = \{v_1,\dots,v_n\}$ and $\mathcal{E} = \{e_1,\dots,e_m\}$, has entries,
\begin{equation}
\label{def_inc}
    b_{i,j} = 
        \begin{cases}
        1, & \text{if } v_i \in e_j, \\
        0, & \text{otherwise.}
        \end{cases}
\end{equation}

A linear algebra definition which will be used in Section \ref{subsec:two_way_effects_hypergraph} is the \textit{face product} between two matrices, described in Definition \ref{def:formal_face_prod}, see also \cite{slyusar1999family}. 

\begin{definition} \label{def:formal_face_prod}
    Let $A$ be an $n \times m$ matrix and $B$ be an $n \times p$ matrix. The face product between $A$ and $B$ is given by the $n \times mp$ matrix, $A \ \square \ B = [a_{ij}B_i]$, where $B_i$ is the $i$th row of matrix $B$. 
\end{definition}

The columns of the face product are given by the element-wise multiplication of pairs of columns such that each column comes from a distinct matrix. An example calculation can be found in \cite{slyusar1999family}.

More generally, the face product of more than two matrices is calculated as pairwise face products working from left to right. For example, $A \ \square \ B \ \square \ C = ( A \ \square \ B ) \ \square \ C$.

The concept of block matrices will also become useful in later sections. A \textit{block matrix} can be partitioned into submatrices called blocks. This is often notated using vertical and horizontal bars, similar to augmented matrices. An example of such a matrix is shown in Example \ref{ex:rest_face}, where the block matrix $A$ is partitioned into three blocks, denoted $A = [ A_1 \vert A_2 \vert A_3 ]$. 

We also introduce the \textit{restricted face power} of a matrix, Definition \ref{def:matrix_prod}, which will be used in Section \ref{sec:Methods}. The notation $A_i$ denotes block $i$ of matrix $A$ and $\square$ is the symbol for the face product, as per Definition \ref{def:formal_face_prod}. 

\begin{definition} \label{def:matrix_prod}
    Let $A=[A_1,\dots,A_p]$ be a matrix with a concordant partition into $p$ blocks. The restricted face power of $A$ to order $n$ is $A^{[n]} = [\underbrace{A_i \ \square \ A_j \ \square \ \cdots \ \square \ A_k}_{n\rm\ blocks} ]$, for all $i < j < \cdots < k$.
\end{definition}

For $n=2$, the restricted face power is a variation of the penetrating face product found in \cite{slyusar1999family}, where the latter takes two matrices and produces the face product between all block pairs such that a pair is comprised from each matrix. Whereas the restricted face power places a strict condition on the pairs of blocks selected, as seen in Definition \ref{def:matrix_prod}. The restrictive condition is illustrated in Example \ref{ex:rest_face}, which shows the restricted face power of order $2$ for an arbitrary matrix $A$.

\begin{example} \label{ex:rest_face}
$$
A = [ A_1 \vert A_2 \vert A_3 ] = 
\left[\begin{array}{@{}cc|cc|cc@{}}
a_{11} & a_{12} & a_{13} & a_{14} & a_{15} & a_{16} \\
a_{21} & a_{22} & a_{23} & a_{24} & a_{25} & a_{26} \\
a_{31} & a_{32} & a_{33} & a_{34} & a_{35} & a_{36} 
\end{array}\right]; 
$$
\resizebox{\textwidth}{!}{$A^{[2]} = \left[
\begin{array}{cccc|cccc|cccc}
a_{11}a_{13} & a_{11}a_{14} & a_{12}a_{13} & a_{12}a_{14} & a_{11}a_{15} & a_{11}a_{16} & a_{12}a_{15} & a_{12}a_{16} & a_{13}a_{15} & a_{13}a_{16} & a_{14}a_{15} & a_{14}a_{16} \\
a_{21}a_{23} & a_{21}a_{24} & a_{22}a_{23} & a_{22}a_{24} & a_{21}a_{25} & a_{21}a_{26} & a_{22}a_{25} & a_{22}a_{26} & a_{23}a_{25} & a_{23}a_{26} & a_{24}a_{25} & a_{24}a_{26} \\
a_{31}a_{33} & a_{31}a_{34} & a_{32}a_{33} & a_{32}a_{34} & a_{31}a_{35} & a_{31}a_{36} & a_{32}a_{35} & a_{32}a_{36} & a_{33}a_{35} & a_{33}a_{36} & a_{34}a_{35} & a_{34}a_{36}
\end{array}\right].$} \\
\end{example}

\section{Methods}
\label{sec:Methods}
In this section we present a novel and general hypergraph based classification algorithm. We assume the training data is stored in an array $F = [f_{i,j}]$, where for $i = 1, \dots, n$ and $j = 1, \dots, m$, entry $f_{i,j}$ records the measurement for unit $i$ and feature $j$. The data is also partitioned into classes, $1,2,\dots,c$, where $Q_i \subseteq \{1,2,\dots,n\}$ denotes the units in class $i$.

In general, features have differing units and scales. As with many classification algorithms, we begin with a normalisation process.

\subsection{Normalisation and Discretisation}
\label{sec:norm_disc}

In the normalisation process all entries of $F$  are first transformed to  standard $z$-scores, as described in \eqref{z_score}, where $\mu_j$ is the mean and $\sigma_j$ is the standard deviation, of column $j$ in $F$,
\begin{equation}\label{z_score}
    z_{i,j} = \frac{f_{i,j} - \mu_j}{\sigma_j}.
\end{equation}
That is, for any feature $j$ the corresponding $z$-scores $\{z_{i,j}\mid i=1,\dots, n\}$ have mean zero and standard deviation one. 

The $z$-score may be discretised via a partition of $\mathbb{R}$,
\begin{equation}\label{partition}
    P_{\ell,\alpha}= \Big\{\Big[ (t-1)\ell + \alpha, t\ell + \alpha\Big ) \mid t \in \mathbb{Z} \Big\}.
\end{equation} 
This partitions the real numbers into intervals, with each interval assigned a unique integer $t$. The parameter $\ell$ is the length of the intervals, and $\alpha$ is a translation factor about the origin.
The score $z_{i,j}$ is discretised by,
\begin{equation} \label{ceil_function}
    d_{i,j} = \Bigg\lceil \frac{z_{i,j} - \alpha}{\ell} \Bigg\rceil.
\end{equation}
Thus $z_{i,j}$ appears in the interval parameterized by $t=d_{i,j}$ in \eqref{partition}. The array $F$ is thus transformed to a normalised and discretised array $D = [d_{i,j}]$ of size $n \times m$. The entries are in the range of integers $\{\min\limits_{i,j}(d_{i,j}),\min\limits_{i,j}(d_{i,j})+1, \dots,  \max\limits_{i,j}(d_{i,j}) \}$ of cardinality
$\tau = \max\limits_{i,j}(d_{i,j}) - \min\limits_{i,j}(d_{i,j}) + 1$.

\subsection{Main Effects Hypergraph Model} 
\label{section:MAIN}

\subsubsection{Model Construction}
\label{sub:model_con}

The array $D$ is used to develop a hypergraph model $\mathcal{H} = (\mathcal{V,E})$, in which each vertex corresponds to a unit $i \in \{1, \dots, n \}$, and the hyperedges are indexed by pairs $(j,t)$, where $j \in \{1, \dots, m\}$ and $t \in \{1, \dots, \tau\}$. Vertex $i$ is incident with hyperedge $(j,t)$ if and only if $d_{i,j} = t$. Thus hyperedges may be grouped by features, and each vertex is incident with $m$ hyperedges, with the possibility that some hyperedges are empty. Notice also that the vertices can be classified into $c$ classes, where $Q_k$ denotes the set of units belonging to class $k$, for $k = 1, \dots, c$. The $n \times \tau m$ incidence matrix for $\mathcal{H}$ is then $B = [b_{i,y}]$, where
\begin{equation}\label{incidence_matrix}
    b_{i,y} = 
        \begin{cases}
        1, & \text{if } y \in \{ \tau(j-1) + d_{i,j} + | \min\limits_{i,j}(d_{i,j})| + 1 \mid  j = 1,\dots,m \}, \\
        0, & \text{otherwise.}
        \end{cases}
\end{equation}

The class incidence of each hyperedge can be summarised in a $c$-tuple, arranged as columns in a $\tau m \times c$ array, $W' = [ w'_{y,k} ]$,
\begin{equation}\label{weight_matrix}
    w'_{y,k} = \frac{1}{|Q_k|}
    \sum\limits_{i\in Q_k} b_{i,y}.
\end{equation}

The rows of $W'$ correspond to hyperedges and columns represent classes. The entries give the proportion of class $k$ units incident with hyperedge $y$.
We normalise each row to generate the {\it hyperedge weight array} $W = [ w_{y,k} ]$,
\begin{equation}\label{normed_weight_matrix}
    w_{y,k} = 
    \begin{cases}
    \frac{w'_{y,k}}{\sum\limits_k w'_{y,k}}, & \sum\limits_k w'_{y,k} \ne 0, \\
    \frac{1}{c}, & \mbox{otherwise.}
    \end{cases}
\end{equation}

 Note that each row sums to one, giving a distribution across classes for each hyperedge. By \eqref{normed_weight_matrix}, empty hyperedges correspond to uniform distributions. 

 This array serves as the model based on the partition $P_{\ell,\alpha}$, denoted $W(P_{\ell,\alpha})$. We can vary the parameters for the partition, giving a population of hyperedge weight arrays, and thus a population of hypergraph models.

\subsubsection{Model Prediction}
\label{subsub:predict}

Given an unclassified unit, with class unknown, we can use the hypergraph model $W(P_{\ell,\alpha})$ to predict the class. An unlabelled unit, $v$, can be represented by the $m$-tuple $v' = (v'_1, \dots, v'_m)$, with each entry recording the measurement of a feature for that unit. We then normalise each entry of $v'$ to standard $z$-scores using \eqref{z_score}, using the respective feature mean and standard deviation calculated from the training data $F$. Next we discretise each normalised entry using $P_{\ell,\alpha}$ and \eqref{ceil_function}. This normalised and discretised $m$-tuple is denoted $v = (v_1, \dots, v_m )$. 
As in \eqref{incidence_matrix}, we convert $v$ to an incidence tuple $\nu = (\nu_1,\dots,\nu_{\tau m})$ 
\begin{equation} \label{incidence_tuple}
    \nu_y = 
    \begin{cases}
        1, & \text{if } y \in \{ \tau(j-1) + v_j + | \min\limits_{i,j}(d_{i,j})| + 1 \mid  j = 1,\dots,m \}, \\
        0, & \text{otherwise.}
        \end{cases}
\end{equation}

We recall that each column of the hyperedge weight array $W(P_{\ell,\alpha})$ defines a probability distribution over the classes for the corresponding hyperedges in $\cal{H}$. We extract the relevant probability distributions with $V = \diag(\nu) \cdot W(P_{\ell,\alpha})$. Let $\overline{\bm{V}}$ be the $c$-tuple of mean values for each column (class) of $V$.
We construct a prediction function  $f: M_{\tau m \times c}(\mathbb{R}) \mapsto \{1,\dots,c\}$
\begin{equation}\label{predict_function}
    f(V) = \argmax\Big(\overline{\bm{V}} \Big).
\end{equation}
Therefore, for a given hypergraph model based on the partition $P_{\ell,\alpha}$, we can use the function $f$ to predict the class of an unlabelled unit. 

\subsubsection{Population of Models and Prediction}
\label{subsub:pop}

We are able to produce a population of models via parameter variation with respect to $\ell$ and $\alpha$. Then for an unlabelled unit, a population of predictions can be made. That is, if we consider $L$ different interval lengths ($\ell$) and $A$ different origins ($\alpha$), then we have $AL$ hypergraph models, thus $AL$ predictions.  

Throughout the following, we select  such pairs ($\ell,\alpha$) randomly from the uniform distribution, $\ell \sim U(0.2,1.5)$ and $\alpha \sim U(-\frac{1}{2},\frac{1}{2})$. The bounds for these distributions were chosen arbitrarily and can be altered at discretion. The number of parameter pairs selected can also vary. 

As such, we let $p_{v}$ be the $AL$-tuple of class predictions across all models for unlabelled unit $v$. We make a final prediction based on the tuple $p_{v}$ according to the function $g: \mathbb{R}^{AL} \mapsto \{1,\dots,c\}$, defined in \eqref{final_function}. The class prediction for the unlabelled unit is given by the function  
\begin{equation}\label{final_function}
    g(p_{v}) = \argmax_k P(p_{v} = k).
\end{equation}

This population of models reduces the variance and provides a more robust final prediction compared to that of a single model, as does a random forest compared to a single decision tree.

\subsection{Two-Way Effects Hypergraph}
\label{subsec:two_way_effects_hypergraph}

The two-way interactions among features, two-way effects, are ignored in Section \ref{section:MAIN}. However, interactions among pairs of features provide further insight into the underlying data \cite{park2004design}. Therefore, the inclusion of such two-way interactions in a classification model may increase predictive performance.

We now construct a hypergraph model which explores two-way interactions. We recall the hypergraph $\cal{H}$ given in Section \ref{sub:model_con}, which is specifically used to study the main effects.
In particular, vertices are units and hyperedges correspond to individual feature and interval pairs. The intersection between two hyperedges of distinct features explores a two-way interaction. As a consequence, we derive a hypergraph based on two-way effects using the hypergraph $\cal{H}$. We define this hypergraph as the \textit{$2$-intersection hypergraph}, as in Definition \ref{def:twoway_hypergraph}.

\begin{definition} \label{def:twoway_hypergraph}
    If $\mathcal{H = (V,E)}$ is a hypergraph where $\mathcal{E} = \{ \epsilon_{j,1},\dots,\epsilon_{j,\tau} \mid j = 1, \dots, m \}$. Then $\mathcal{H}_2 = (\mathcal{V},\mathcal{E}_2)$ is the $2$-intersection hypergraph of $\mathcal{H}$, where $\mathcal{E}_2 = \{ \epsilon_{j,t} \cap \epsilon_{\hat{j},\hat{t}} \mid j \neq \hat{j} \ \land \ t,\hat{t} \in \{1, \dots, \tau \} \} $.
\end{definition}

The number of distinct pairs of hyperedges is the product of the number of distinct pairs of features $\binom{m}{2}$ with the number of distinct feature pairs, $\tau^2$. Therefore $| \mathcal{E}_2 | = \binom{m}{2}\tau^2$, with some hyperedges possibly being empty. 

In order to follow Algorithm \ref{subsub:alg} using a $2$-intersection hypergraph, we require an incidence matrix representation. This incidence matrix can be derived from the original hypergraph incidence matrix using the restricted face power as per Definition \ref{def:matrix_prod}. Therefore, the incidence matrix of a $2$-intersection hypergraph is described in Definition \ref{def:2-inc}. We can use this incidence matrix and Algorithm \ref{subsub:alg} to formulate a hypergraph model based on two-way effects, thus classifying unlabelled units.

\begin{definition} \label{def:2-inc}
    If $\mathcal{H}$ is a hypergraph with incidence matrix $B$, then the incidence matrix of the $2$-intersection hypergraph $\mathcal{H}_2$ is $B^{[2]}$.
\end{definition}

\subsection{$\bm{\eta}$-Way Effects Hypergraph}

The hypergraph models described thus far consider the one and two-way effects among features. However, higher order interactions have the potential to illuminate detail on the dataset which may otherwise have been missed \cite{varela2006analysis}. Therefore, we give the definitions for a general hypergraph model which explore $\eta$-way interactions for $\eta \geqslant 3$. We begin with the $\eta$-intersection hypergraph, as found in Definition \ref{def:n-int_hyp}.

\begin{definition} \label{def:n-int_hyp}
    If $\mathcal{H = (V,E)}$ is a hypergraph where $\mathcal{E} = \{ \epsilon_{j,1},\dots,\epsilon_{j,\tau} \mid j = 1, \dots, m \}$. Then $\mathcal{H}_n = (\mathcal{V},\mathcal{E}_n)$ is the $\eta$-intersection hypergraph of $\mathcal{H}$, where $\mathcal{E}_n = \{ \epsilon_{j,t} \cap \cdots \cap \epsilon_{\hat{j},\hat{t}} \mid | \{j,\dots,\hat{j} \}| = n \ \land \ j \neq \cdots \neq \hat{j} \ \land \ t,\dots,\hat{t} \in \{1,\dots,\tau\}   \} $.
\end{definition}

We now use the restricted face power (Definition \ref{def:matrix_prod}) to construct the incidence matrix for the $\eta$-intersection hypergraph as per Definition \ref{def:n_face_prod}.

\begin{definition} \label{def:n_face_prod}
    If $\mathcal{H}$ is a hypergraph with incidence matrix $B$, then the incidence matrix of the $\eta$-intersection hypergraph $\mathcal{H}_\eta$ is $B^{[\eta]}$.
\end{definition}

With this, the incidence matrix construction for the $\eta$-intersection hypergraph can be found in Definition \ref{def:n_face_prod}. This incidence matrix can be applied in the construction of $W(P_{\ell,\alpha})$ in Step 5 of
 Algorithm \ref{subsub:alg} to obtain a hypergraph model based on $\eta$-way interactions. Subsequently, we define the $\eta$-way effects hypergraph model as the $\mathcal{H}_\eta$-algorithm. 

\section{Case Studies}
\subsection{Benchmark Classification Algorithms}

In the following case studies, we compare the performance of a benchmark classification algorithm and the hypergraph algorithm (Algorithm \ref{subsub:alg}), using the $5$-fold cross-validation accuracy.  The benchmark classification algorithm we consider is random forests, a robust and accurate classification tool, see \cite{breiman2001random,diaz2006gene}. We use the generic ensemble function with bagged trees which can be found in MATLAB \cite{MATLAB}. 

\subsection{Fisher's Iris Dataset}

Fisher's Iris dataset \cite{fisher1936use} is well-studied within the fields of machine learning and statistics and commonly employed as a benchmark for assessing classification algorithms \cite{mahesh2020machine}. This is attributed to its straightforward nature and clearly defined classes. Specifically, the dataset encompasses measurements of four distinct features taken from three species of iris flowers; specific details are shown in Tables \ref{tab:fisher_species} and \ref{tab:fisher_features}.

\begin{table}[h!]
\parbox{.45\linewidth}{
\centering
\begin{tabular}{|c|c|c|}
\hline
\textbf{Class} &  \textbf{Species} & \textbf{Units}  \\
\hline
$1$ & \textit{Iris Setosa} & $50$  \\
\hline
$2$ & \textit{Iris Virginica} & $50$ \\
\hline
$3$ & \textit{Iris Versicolor} & $50$ \\
\hline
\end{tabular}
\caption{List of species in Fisher's Iris dataset.}
\label{tab:fisher_species}
}
\hfill
\parbox{.45\linewidth}{
\centering
\begin{tabular}{|c|c|}
\hline
\textbf{Feature Number} &  \textbf{Feature} \\
\hline
$1$ & \textit{Sepal Length} \\
\hline 
$2$ & \textit{Sepal Width} \\
\hline
$3$ & \textit{Petal Length} \\
\hline
$4$ & \textit{Petal Width} \\
\hline
\end{tabular}
\caption{List of features in Fisher's Iris dataset.}
\label{tab:fisher_features}
}
\end{table}

We now evaluate the performance of the hypergraph based approach with respect to this dataset, as well as the benchmark classification algorithm, random forests.  Since this particular dataset only contains four features, we only consider the $\mathcal{H}_{\eta}$-algorithm for $\eta = 1,2$. The accuracy scores can be found in Table \ref{tab:iris_res}, along with the number of parameter variations for each $\mathcal{H}_{\eta}$-algorithm. That is, $L$ and $A$ give the number of different interval lengths and origins respectively. 

\begin{table}[h!]
\centering
\begin{tabular}{|c|c|c|c|}
\hline
\textbf{Algorithm} & \textbf{Accuracy} & \textbf{Standard Error} & \textbf{Parameter Selection}  \\
\hline
Random Forest & $0.9443$ & 0.0188 & $-$ \\
\hline
$\mathcal{H}_1$-Algorithm & $0.9514$ & 0.0187 & $L = 200, A = 10$  \\
\hline
$\mathcal{H}_2$-Algorithm & $0.9431$ & 0.0237 & $L = 200, A = 10$  \\
\hline
\end{tabular}
\caption{The $5$-fold cross-validation accuracy scores of algorithms on Fisher's Iris dataset.}
\label{tab:iris_res}
\end{table}

The results in Table \ref{tab:iris_res} suggest that all models perform similarly. Specifically, we see the $\mathcal{H}_1$-algorithm achieving the greatest accuracy, followed closely by the random forest and $\mathcal{H}_2$-algorithm respectively. In general, we expect to see the $\mathcal{H}_2$-algorithm outperforming the $\mathcal{H}_1$-algorithm. However, this is not the case for the Iris dataset. This result may be explained by the limited number of features (four), with the main effects better partitioning the classes compared to the two-way interactions. That is, the two-way interactions introduce noise to the classifier which decreases performance. 

We note here that the literature contains finely tuned random forest algorithms which achieve greater accuracy on this dataset compared to the results shown here. However, in this study we did not want to over-customise for this particular dataset. Nevertheless, these results suggest the hypergraph models are capable of producing similar accuracy compared to a standard benchmark classification algorithm. 

\subsection{Starch Grain Dataset}

We now evaluate the performance of the hypergraph models on a more complex dataset, that encapsulates representative information for starch grains from seven distinct species. In this setting the model is being proposed as a technique for use in identifying plant micro-fossils of unknown species origin through comparison with reference grains. Starch grains grow in the spaces between the cells in plants. Initially growth is unconstrained, with immature grains being small and spherical. Over time space and crowding induce growth patterns more representative of the individual species. While it is not possible to characterise starch grains by maturity, through two-dimensional projections each starch grain can be characterised by shape metrics and supplemented by associated Fourier signatures. The methodology of extraction for the starch grain features can be found in \cite{coster2015starch}, with the particular dataset studied in this paper given as personal communication from the authors of \cite{coster2015starch}. 

The two-dimensional projections introduce variance in the data due to the orientation of the grains. As a result, we expect the performance of classifiers on this set of data to be reduced in comparison to Fisher's Iris dataset. This is due to the overlapping values for the different features, as shown in \cite{coster2015starch}, making classification on this dataset to be a difficult problem. In total there are $944$ units (grains) from seven distinct species, with each unit described by $16$ morphological features (shape metrics and the Fourier signature), as summarised in Tables \ref{tab:grain_species} and \ref{tab:grain_features}.

\begin{table}[h!]
\parbox{.45\linewidth}{
\centering
\resizebox{0.5\textwidth}{!}{\begin{tabular}{|c|c|c|}
\hline
\textbf{Class} &  \textbf{Species} & \textbf{Units}  \\
\hline
$1$ & \textit{Dysphania kalpari (DK)} & $123$ \\
\hline
$2$ & \textit{Acacia aneura (AA)} & $105$ \\
\hline
$3$ & \textit{Acacia victoriae (AV)} & $121$ \\
\hline
$4$ & \textit{Brachiaria miliiformis (BM)} & $134$ \\
\hline
$5$ & \textit{Eragrostis eriopoda (EE)} & $154$ \\
\hline
$6$ & \textit{Yakirra australiensis (YA)} & $142$ \\
\hline
$7$ & \textit{Brachychiton populneus (BP)} & $165$ \\
\hline
\end{tabular}}
\caption{List of species in starch grain dataset.}
\label{tab:grain_species}
}
\hfill
\parbox{.5\linewidth}{
\centering
\resizebox{0.45\textwidth}{!}{\begin{tabular}{|c|c|}
\hline
\textbf{Feature Number} &  \textbf{Feature}  \\
\hline
$1$ & \textit{Length}  \\
\hline
$2$ & \textit{Area} \\
\hline
$3$ & \textit{Perimeter} \\
\hline
$4$ & \textit{Circularity}  \\
\hline
$5$ & \textit{Hilum Position}  \\
\hline
$6$ & \textit{Fourier Coefficient $0$}  \\
\hline
$\vdots$ & $\vdots$  \\
\hline
$16$ & \textit{Fourier Coefficient $10$}  \\
\hline
\end{tabular}}
\caption{List of features in starch grain dataset.}
\label{tab:grain_features}
}
\end{table}

\subsubsection{Classification Results}

We now apply the $\mathcal{H}_{\eta}$-algorithm for $\eta = 1,2,3$ to the starch grain dataset, with $L = 80$ and $A = 25$ to achieve a population of $2000$ hypergraph models. The random forest algorithm will be used as a benchmark classifier. We use $5$-fold cross-validation, where the units for each species were separately partitioned into five sets of approximately equal size. The detail of this process is provided in Table \ref{tab:5fold_values} in the Appendix. The $5$-fold cross-validation accuracy scores for the respective algorithms can be found in Table \ref{tab:grain_res}. 

\begin{table}[h!]
\centering
\begin{tabular}{|c|c|c|c|}
\hline
\textbf{Algorithm} & \textbf{Accuracy} & \textbf{Standard Error} & \textbf{Parameter Selection}  \\
\hline
Random Forest & 0.5471 & 0.0092 & $-$ \\
\hline
$\mathcal{H}_1$-Algorithm & 0.5225 & 0.0098 & $L = 80, A = 25$  \\
\hline
$\mathcal{H}_2$-Algorithm & 0.5578 & 0.0090 & $L = 80, A = 25$  \\
\hline
$\mathcal{H}_3$-Algorithm & 0.5601 & 0.0097 & $L = 80, A = 25$  \\
\hline
\end{tabular}
\caption{The $5$-fold cross-validation accuracy scores of algorithms on the starch grain dataset.}
\label{tab:grain_res}
\end{table}

Table \ref{tab:grain_res} indicates that for $\eta = 2,3$, the $\mathcal{H}_{\eta}$-algorithm shows improved accuracy over the random forest, whereas the random forest outperforms the case where $\eta=1$. Therefore, increasing $\eta$ in the $\mathcal{H}_{\eta}$-algorithm coincides with increased accuracy, demonstrating that incorporating multi-way interactions provides salient information in the dataset which is otherwise overlooked. 

\begin{figure}[h!]
    \centering
    \includegraphics[width=0.5\textwidth]{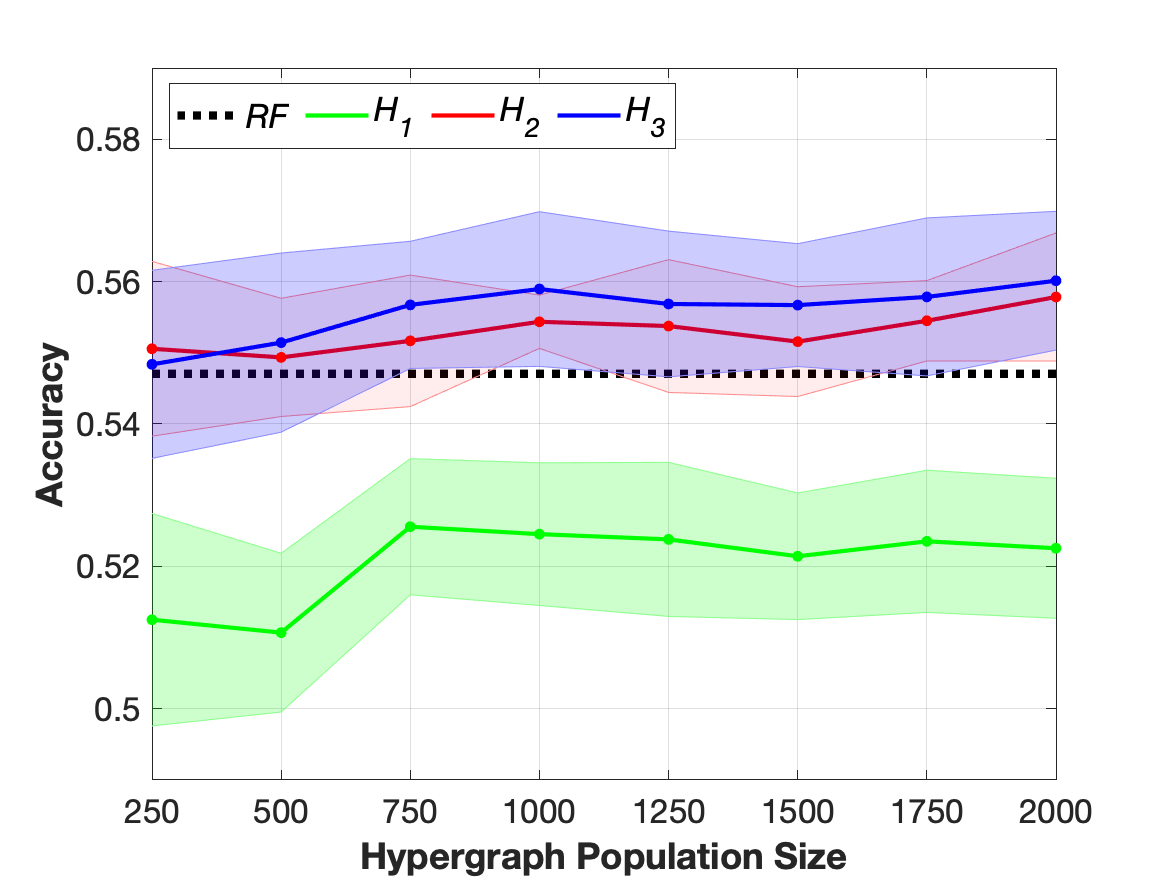}
    \caption{Accuracy scores across different hypergraph population sizes.}
    \label{fig:pop_results}
\end{figure}

Along with the results found in Table \ref{tab:grain_res}, we also explore the impact of hypergraph population size and the performance of the $\mathcal{H}_{\eta}$-algorithm. That is, in Figure \ref{fig:pop_results}, we show the accuracy of the $\mathcal{H}_{\eta}$-algorithm across various sizes of the hypergraph population. We indicate the standard error as the shaded regions for each algorithm. 

As illustrated in Figure \ref{fig:pop_results}, the accuracy is increased as the number of models in the population increases from $250$ to $2000$, across all $\eta \in \{1,2,3\}$. This shows in general, a greater selection of hypergraph models will increase prediction performance. However, we also notice that for small population sizes ($\leqslant 1000$), the accuracy scores fluctuate.

\subsubsection{Detailed Classification Performance}

To explore the salient properties of the $\mathcal{H}_{\eta}$-algorithm, we provide and discuss detailed performance metrics. This discussion evaluates the strengths and weaknesses specific to this case study but also illuminates how the algorithm can be optimised in a general setting. The detailed results for the the $\mathcal{H}_3$-algorithm and random forest are shown in Table \ref{tab:conf_RFvsH3}, with the respective confusion matrices given in Tables \ref{tab:Conf_H3} and \ref{tab:Conf_RF} in the Appendix. 

\begin{table}[h!]
\settowidth\mylenA{False Negative Rate} 
\setlength\mylenB{(\mylenA-2\tabcolsep)/2}
    \centering
    \resizebox{\textwidth}{!}{\begin{tabular}{| c | *{2}{wc{\mylenB}|} *{2}{wc{\mylenB}|} *{2}{wc{\mylenB}|} *{2}{wc{\mylenB}|} }
    \hline
    \multirow{2}{*}{\textbf{Species}} & 
    \multicolumn{2}{c|}{\textbf{False Negative Rate}} &
    \multicolumn{2}{c|}{\textbf{False Positive Rate}} &
    \multicolumn{2}{c|}{\textbf{True Positive Rate}} &
    \multicolumn{2}{c|}{\textbf{True Negative Rate}} \\ 
    \cline{2-9}
    & $RF$ & $\mathcal{H}_3$ & $RF$ & $\mathcal{H}_3$ & $RF$ & $\mathcal{H}_3$ & $RF$ & $\mathcal{H}_3$ \\ \hline
    \textit{Dysphania kalpari} & 16.1\%	& 7.3\%	& 6.4\%	& 8.9\%	& 83.9\%	& 92.7\% & 93.6\% & 91.1\% \\ \hline 
    \textit{Acacia aneura} & 45.0\% & 21.0\% & 6.0\% & 7.7\% & 55.0\% & 	79.0\% & 94.0\% & 92.3\%  \\ \hline 
    \textit{Acacia victoriae} & 62.6\% & 53.7\%	& 7.9\%	& 8.9\%	& 37.4\% & 46.3\%	& 92.1\% & 91.1\% \\ \hline 
    \textit{Brachiaria miliiformis} & 48.5\% & 41.0\% &	9.7\% &	10.5\%	& 51.5\% & 59.0\% &	90.3\% & 89.5\% \\ \hline 
    \textit{Eragrostis eriopoda} & 57.7\% & 71.4\% & 8.4\% & 5.1\% & 	42.3\% & 28.6\% & 91.6\% & 94.9\%  \\ \hline 
    \textit{Yakirra australiensis} & 44.4\% & 52.8\% & 6.6\% & 	5.1\% & 55.6\% & 47.2\% & 93.4\% & 	94.9\%  \\ \hline 
    \textit{Brachychiton populneus} & 43.7\% & 48.5\% & 8.4\% & 	5.0\% & 56.3\% & 51.5\% & 91.6\% & 95.0\%  \\ \hline 
    \end{tabular}}
\caption{True and false positive and negative rates for the $\mathcal{H}_3$-algorithm.}
\label{tab:conf_RFvsH3}
\end{table}

The true positive rate per class, which is the ratio between the number of correct classifications and size of the true class, varies among species. In general, the $\mathcal{H}_3$-algorithm outperforms the random forest with respect to the true positive rates for species $DK$, $AA$, $AV$ and $BM$. Both classifiers achieve a true positive rate greater than $80\%$ when classifying the $DK$ species. Since this performance is not seen for the other species, we evaluate the $DK$ units to be most distinct with respect to the features studied. The worst performance within the $\mathcal{H}_3$-algorithm was for $EE$, and $AV$ for the random forest. The $\mathcal{H}_3$-algorithm confusion matrix (Table \ref{tab:Conf_H3}) found in Appendix A indicates that this algorithm most commonly confuses $EE$ with $DK$ and $BM$. The pattern of confusing $EE$ with $BM$ was also found in the tree classifier constructed in \cite{coster2015starch}. This confusion was explained due to an overlap in features, with $EE$ and $BM$ having similar Fourier signatures for higher harmonics \cite{coster2015starch}. After considering the distributions of the geometric morphometric measures found in \cite{coster2015starch}, the confusion between $EE$ and $DK$ could be due to the similarity in the \textit{Circularity} feature. Specifically, the three species, $EE$, $DK$ and $YA$, have similar \textit{Circularity} distributions which may lead the algorithm to predict $EE$ units as any one of these species, a problem that would be compounded through the presence of immature starch grains. See Section \ref{subsc:ruleout} for further discussion on this point.

\subsubsection{Threshold Classification Performance}
\label{subsub:threshold_classification_performance}

\begin{figure}[h!]
    \centering
    \includegraphics[width=0.5\textwidth]{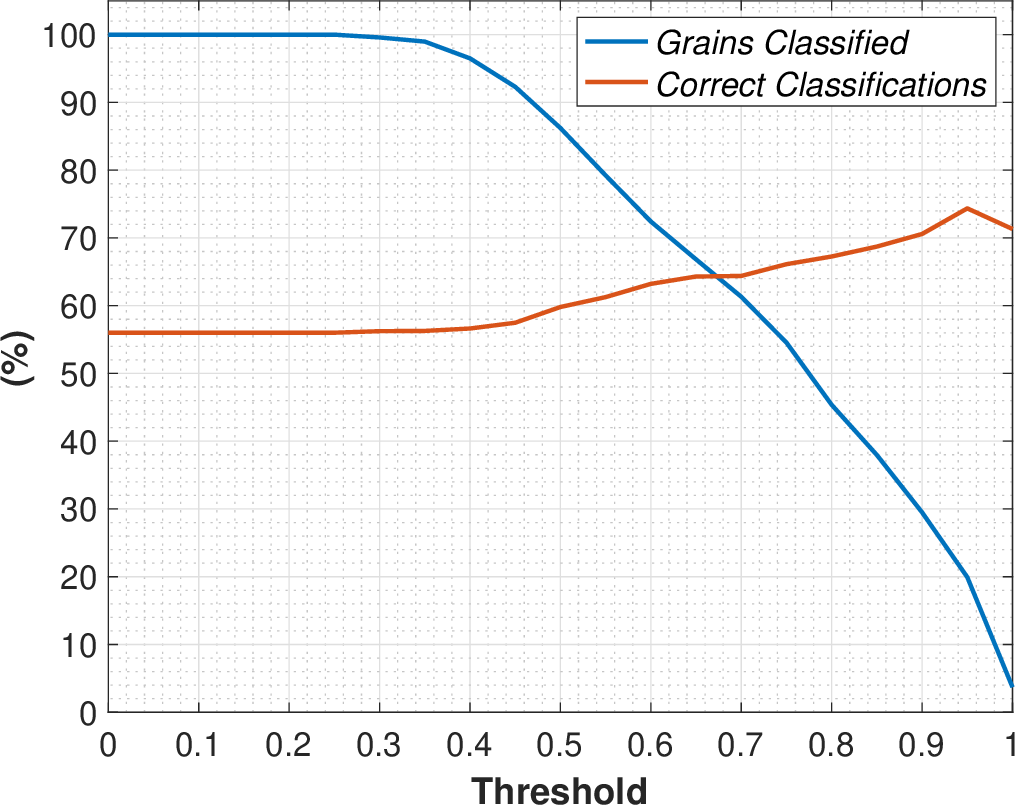}
    \caption{The accuracy of the $\mathcal{H}_3$-algorithm  (red) and the percentage of units classified (blue) across decision threshold values.}
    \label{fig:threshold_res}
\end{figure}

The performance of the $\mathcal{H}_3$-algorithm can be improved by introducing a \textit{decision threshold}. In classification algorithms, a decision threshold is a pre-determined value which is the minimum requirement for a prediction to be accepted, otherwise the prediction is unclassified. The aim is to minimise false negative and positive rates, while increasing true positive and negative rates. As shown in Figure \ref{fig:threshold_res}, the drawback of a decision threshold is the introduction of unclassified units. These are units which cannot be classified and therefore need further investigation. Moreover, with an increased threshold value comes an increased number of unclassified units. 

Standard classification algorithms often use a threshold of $0.5$, however this arbitrary value is shown to be inappropriate in some cases \cite{chen2006decision}. There are various techniques for finding the best decision threshold, see for example \cite{zou2016finding}. In \cite{coster2015starch} a threshold of $0.75$ was used; we will use the same threshold as we are studying the same dataset. 

To incorporate a decision threshold into Algorithm \ref{subsub:alg}, we recall that $p_v$, for a given unit ($v$) in the validation set, is a tuple storing the class prediction for each model. The class which appears most often is the final predicted class for that unit. It is here that we introduce the decision threshold. For class $k$ to be the final prediction of an unit, we require $P(p_v = k) > 0.75$; otherwise the unit is not classified. This will ensure that at least $75\%$ of the models agree that class $k$ is the desired prediction. We incorporate this in Algorithm \ref{subsub:alg} as \textbf{Output($\star$)}. The detailed results of the confusion matrix for the $\mathcal{H}_3$-algorithm with and without the threshold are shown in Table \ref{tab:conf_res_threshold}.

\begin{table}[th]
\settowidth\mylenA{False Negative Rate} 
\setlength\mylenB{(\mylenA-2\tabcolsep)/2}
    \centering
    \resizebox{\textwidth}{!}{\begin{tabular}{| c | *{2}{wc{\mylenB}|} *{2}{wc{\mylenB}|} *{2}{wc{\mylenB}|} *{2}{wc{\mylenB}|} }
    \hline
    \multirow{2}{*}{\textbf{Species}} & 
    \multicolumn{2}{c|}{\textbf{False Negative Rate}} &
    \multicolumn{2}{c|}{\textbf{False Positive Rate}} &
    \multicolumn{2}{c|}{\textbf{True Positive Rate}} &
    \multicolumn{2}{c|}{\textbf{True Negative Rate}} \\ 
    \cline{2-9}
    & $\mathcal{H}_3(0.75)$ & $\mathcal{H}_3(0)$ & $\mathcal{H}_3(0.75)$ & $\mathcal{H}_3(0)$ & $\mathcal{H}_3(0.75)$ & $\mathcal{H}_3(0)$ & $\mathcal{H}_3(0.75)$ & $\mathcal{H}_3(0)$ \\ \hline
    \textit{Dysphania kalpari} & 5.0\% & 7.3\% & 11.6\%	& 8.9\%	& 95.0\%	& 92.7\%	& 88.4\% & 91.1\% \\ \hline 
    \textit{Acacia aneura} & 7.6\% &	21.0\% &	8.2\% &	7.7\% &	92.4\% &	79.0\% &	91.8\% &	92.3\%  \\ \hline 
    \textit{Acacia victoriae} & 55.2\% &	53.7\% &	5.5\% &	8.9\% &	44.8\% &	46.3\% &	94.5\% &	91.1\% \\ \hline 
    \textit{Brachiaria miliiformis} & 23.3\% &	41.0\% &	7.7\% &	10.5\% &	76.7\% &	59.0\% &	92.3\% &	89.5\% \\ \hline 
    \textit{Eragrostis eriopoda} & 77.6\% &	71.4\% &	1.3\% &	5.1\% &	22.4\% &	28.6\% &	98.7\% &	94.9\%  \\ \hline 
    \textit{Yakirra australiensis} & 40.3\% &	52.8\% &	0.9\% &	5.1\% &	59.7\% &	47.2\% &	99.1\% &	94.9\%  \\ \hline 
    \textit{Brachychiton populneus} & 41.6\% &	48.5\% &	4.8\% &	5.0\% &	58.4\% &	51.5\% &	95.2\% &	95.0\%  \\ \hline 
    \end{tabular}}
\caption{True and false positive and negative rates for the $\mathcal{H}_3$-algorithm with ($\mathcal{H}_3(0.75)$) and without ($\mathcal{H}_3(0)$) the decision threshold.}
\label{tab:conf_res_threshold}
\end{table}

This decision threshold predicts $54.6\%$ of the units in the validation set with an overall accuracy of $66.1\%$. This is an accuracy increase of approximately $10\%$. The true positive are seen to increase in all species except $AV$ and $EE$, with both having a small decrease. Further investigation into the confusion matrix, found in Table \ref{tab:Conf_H3_threshold} (Appendix A), reveals that even with the decision threshold the $EE$ species still gets confused with the $DK$ and $BM$ species. This demonstrates that the algorithm is unable to distinguish between the similar features for this species. While $AV$ is often confused with $AA$ and $BP$, which also occurs in the $\mathcal{H}_3$-algorithm without a decision threshold and random forest (Tables \ref{tab:Conf_H3} and \ref{tab:Conf_RF}). The confusion between $AV$ and $BP$ can be explained by the MANOVA analysis conducted in \cite{coster2015starch}, which shows that $AV$ and $BP$ are closely related with respect to Mahalanobis distance. While the confusion between $AV$ and $AA$ is a result of the similar \textit{Circularity} distribution as commented in \cite{coster2015starch}. Further investigation of the \textit{Circularity} distribution reveals that $AV$, $AA$ and $EE$ all have similar distributions which may cause confusion when classifying these species. The species $DK$, $EE$ and $YA$ also have similar \textit{Circularity} distributions which may cause confusion, as noted earlier and discussed further in the next section.

\subsubsection{Threshold Classification to Rule Out Species}\label{subsc:ruleout}

Classifiers predict a particular outcome based on the patterns of known data. However, we can also consider narrowing down the possible outcomes by ruling some out. That is, we want to exchange less specific predictions for an improvement in accuracy. In reference to the starch grain dataset, we look to remove a set of species with certainty for the prediction of an unknown unit. As discussed earlier, the dataset itself is the two-dimensional projection of the starch grain shapes with the starch grains growing in the spaces between the cells in plants. Initially, they are not constrained so they are all spherical, and separating the early stage grains between species is nigh on impossible. When the space becomes more crowded though, the grains tend to take on characteristic shapes due to the packing. Therefore, the general problem of taking an unknown starch grain and trying to identify it by comparing it to known samples is difficult. This is because we cannot be sure that we have the correct comparator samples and there is no guarantee that the species can be distinguished from each other. As such, we now aim to use the information given in the $\mathcal{H}_{\eta}$-algorithm to narrow the possibilities down to a subset of the species; identifying those subsets that cannot be separated. 

We will consider this problem using two different techniques, both utilising different aspects of the $\mathcal{H}_{\eta}$-algorithm. The first technique will use $p_v$ (defined in Section \ref{subsub:pop}) while the second technique will use $\overline{\bm{V}}$ (defined in Section \ref{subsub:predict}). These two techniques use different information from the $\mathcal{H}_{\eta}$-algorithm, with the latter using the probability distributions encoded by the hyperedges, while the former uses the class predictions arising from these distributions. In both cases, a threshold will be introduced to rule out species. We will evaluate the performance using accuracy and also the number of species ruled out. 

We begin by discussing the technique which involves $p_v$. We recall that for an unlabelled unit $v$, the tuple $p_v$ stores the class predictions across all models, where $AL$ is the total number of model predictions. The normalised frequency of each class prediction is then stored in the tuple $N(p_v) = (n_1,\dots,n_c)$ where entries are defined as, $$n_i = \frac{\sum\limits_{x \in p_v}\mathbf{1}_{x = i}}{AL}.$$ 

In order to rule out a set of species we introduce a threshold, $\alpha$, where we only consider the species where the cumulative frequency is at least $\alpha$. This is described formally in Definition \ref{def:tech1}.

\begin{definition} \label{def:tech1}
    Given the normalised frequency tuple $N(p_v) = (n_1,\dots,n_c)$ and threshold $\alpha$, we select the set of species for the unlabeled unit $v$ as, $$P_v(\alpha,N(p_v)) = \min \{ i \mid \sum n_i \geqslant \alpha \}.$$
\end{definition}

\begin{figure}[h!]
\centering
\begin{minipage}{0.45\textwidth}
    \centering
    \includegraphics[width=1\textwidth]{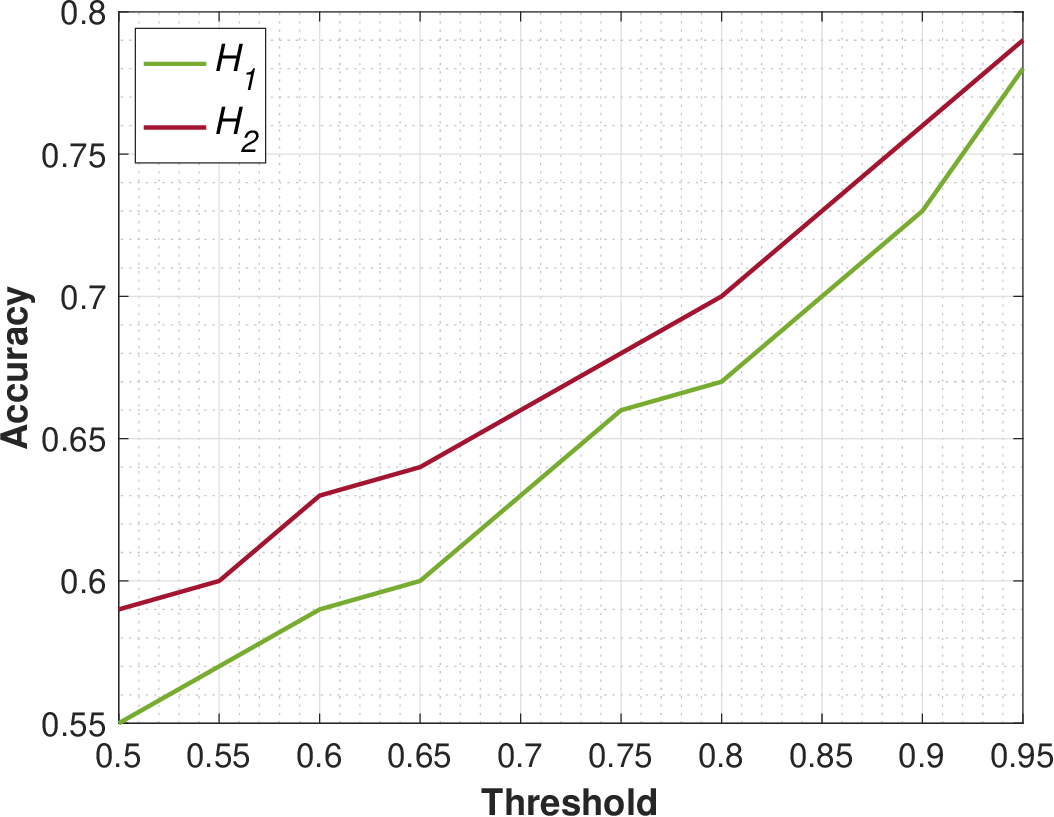}
    \caption{The accuracy scores achieved for different threshold values when using the class prediction technique for the $\mathcal{H}_1$ and $\mathcal{H}_2$-algorithms.}
    \label{fig:tech1_acc}
\end{minipage}\hfill
\begin{minipage}{0.45\textwidth}
    \centering
    \includegraphics[width=1\textwidth]{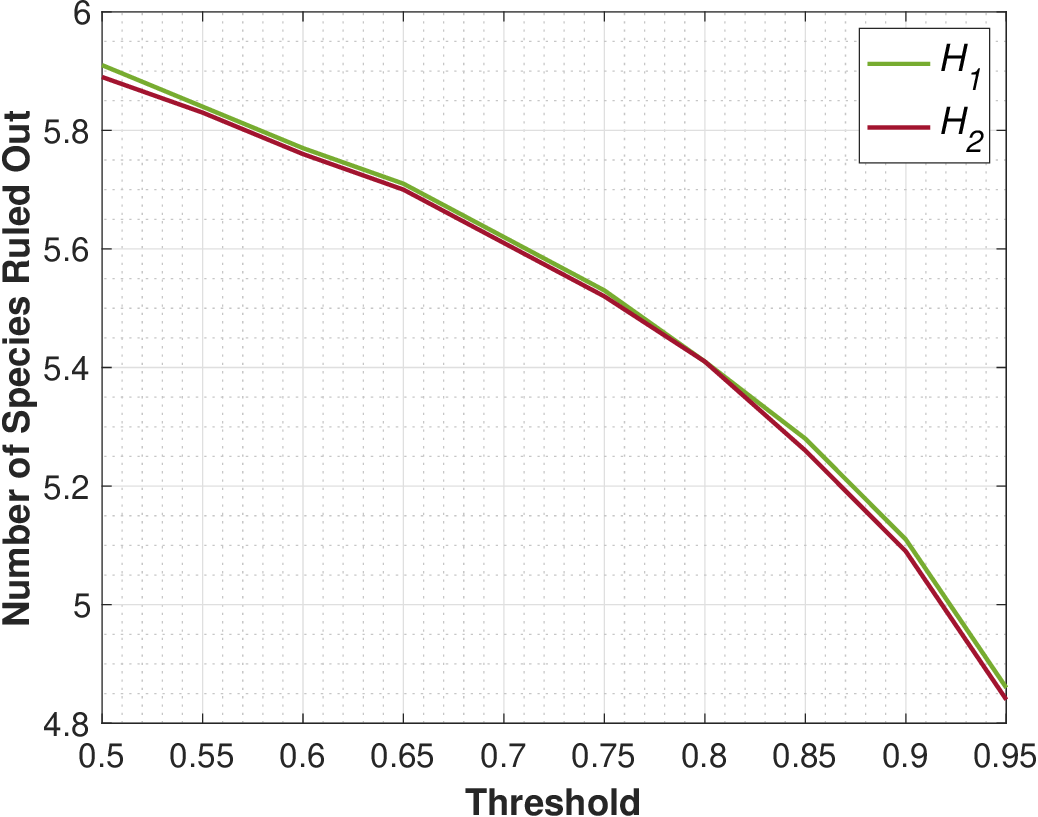}
    \caption{The number of species ruled out for different threshold values when using the class prediction technique for the $\mathcal{H}_1$ and $\mathcal{H}_2$-algorithms.}
    \label{fig:tech1_rule}
\end{minipage}
\end{figure}

With this definition, we look to only consider species for which the majority of class predictions agree. We now evaluate the performance of this technique on accuracy, where for an unlabelled unit the technique is successful if the true species is in the set of predicted species. Figure \ref{fig:tech1_acc} shows the accuracy across different threshold values for the $\mathcal{H}_1$ and $\mathcal{H}_2$-algorithm. In particular, we notice that accuracy increases with larger threshold values. We also see a noticeable difference between the algorithms, with the $\mathcal{H}_2$-algorithm outperforming the $\mathcal{H}_1$-algorithm. 

Along with accuracy as an evaluation of performance, we also want to consider how many species we rule out. This information can be found in Figure \ref{fig:tech1_rule}, which shows the average number of species each algorithm was able to rule out across different threshold values. We see very similar behaviour in this respect for both the $\mathcal{H}_1$ and $\mathcal{H}_2$-algorithm. This result and Figure \ref{fig:tech1_acc} reconfirm that the higher order interactions captured by the hypergraph structure improve performance. Moreover, Figure \ref{fig:tech1_rule} reveals that all threshold values are able to rule out over half of the species. Specifically, threshold values less than $0.9$ rule out more than five species on average. Overall, this technique shows an improved accuracy when compared to earlier results while ruling out over half of the species on average. However, since we are only ruling out species, we desire an increased performance with respect to accuracy. 

We now consider the second technique which makes use of the $\overline{\bm{V}}$ tuple. We recall that for a given hypergraph model and unlabelled unit, $\overline{\bm{V}}$ stores the mean across the probability distributions given by the hyperedges for each species. In this technique, we will use the $AL$ tuples to rule out species. Going forward, we let $\overline{\bm{V}}_{\pi}$ be the tuple associated with model $\pi$, where $\pi = 1, \dots, AL$. Since we have $AL$ tuples, we find the element-wise mean across all these tuples and let this be denoted as $\overline{\rho} = (\rho_1,\dots,\rho_c)$. With this, we can now introduce a threshold similar to the previous technique, as per Definition \ref{def:tech2}.

\begin{definition} \label{def:tech2}
    Given the element-wise mean tuple $\overline{\rho} = (\rho_1,\dots,\rho_c)$ and threshold $\alpha$, we select the set of species for the unlabeled unit $v$ as, $$P_v(\overline{\rho},\alpha) = \min \{ i \mid \sum \rho_i \geqslant \alpha \}.$$
\end{definition}

\begin{figure}[h!]
\centering
\begin{minipage}{0.45\textwidth}
    \centering
    \includegraphics[width=1\textwidth]{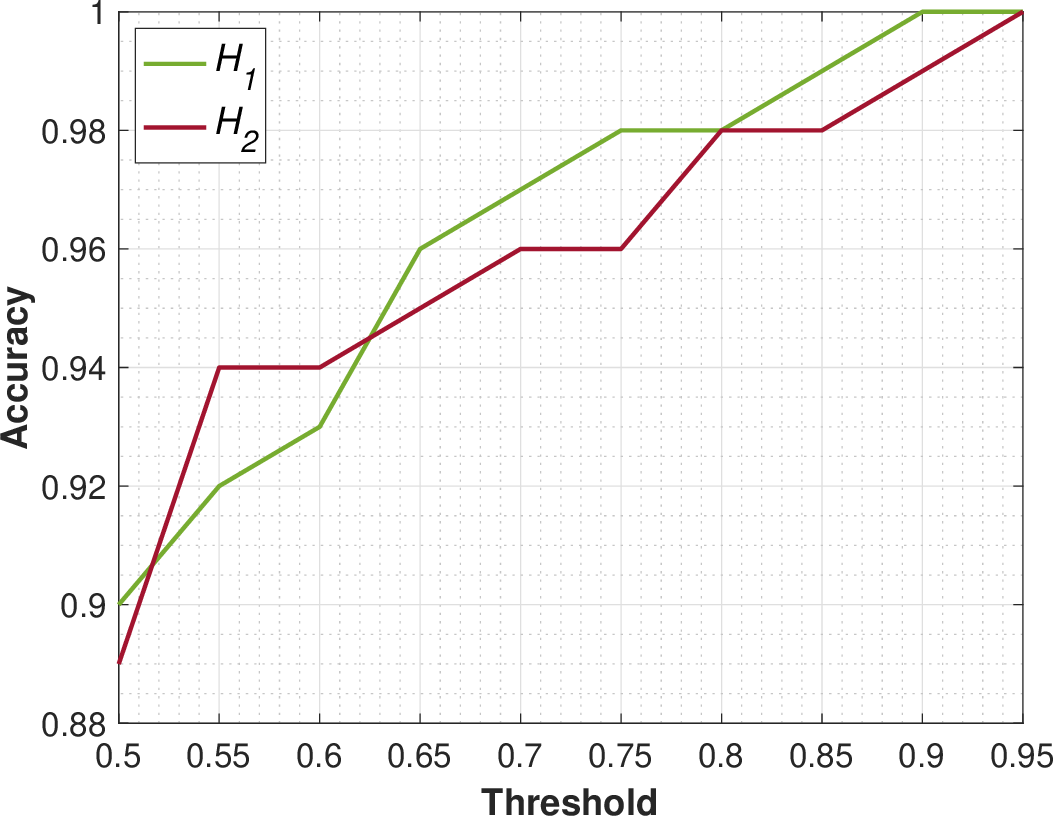}
    \caption{The accuracy scores achieved for different threshold values when using the probability distribution technique for the $\mathcal{H}_1$ and $\mathcal{H}_2$-algorithms.}
    \label{fig:tech2_acc}
\end{minipage}\hfill
\begin{minipage}{0.45\textwidth}
    \centering
    \includegraphics[width=1\textwidth]{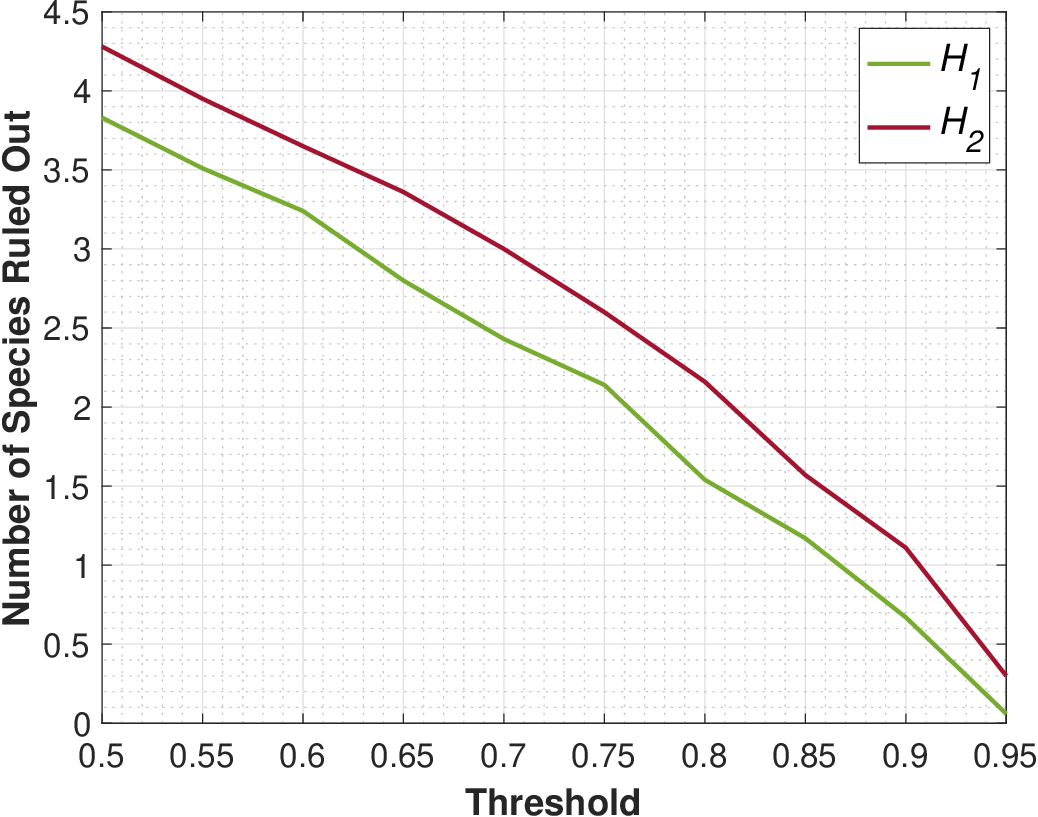}
    \caption{The number of species ruled out for different threshold values when using the probability distribution technique for the $\mathcal{H}_1$ and $\mathcal{H}_2$-algorithms.}
    \label{fig:tech2_rule}
\end{minipage}
\end{figure}

With this definition, we can now analyse the performance of this technique with respect to accuracy and the number of species ruled out. These results are shown in Figures \ref{fig:tech2_acc} and \ref{fig:tech2_rule}. In comparison to Figure \ref{fig:tech1_acc}, we see that both algorithms have a much higher accuracy. However, the results shown in Figure \ref{fig:tech2_rule} reveal that the upper range of threshold values rule out minimal species; for example, threshold values greater than $0.9$ rule out less than one species on average for both algorithms. 

Overall, we see that both techniques described here were successful in ruling out sets of species. In both cases we saw an increase in accuracy, compared to the results given in Figure \ref{fig:pop_results}, which was expected since we are classifying for a set of species rather than a single species. To compare techniques across algorithms, we consider what threshold is required to achieve $90\%$ and $95\%$ accuracy, as seen in Tables \ref{tab:90_acc} and \ref{tab:95_acc} respectively. We compare the technique and algorithm combinations with respect to the exchange between accuracy and the number of species ruled out. 

\begin{table}[h!]
\centering
\begin{tabular}{|c|c|c|c|}
\hline
\textbf{Technique} & \textbf{Algorithm} & \textbf{Threshold} & \textbf{Number of Species Ruled Out}  \\
\hline
Class Prediction & $\mathcal{H}_1$-Algorithm & $-$ & $-$ \\
\hline
Class Prediction & $\mathcal{H}_2$-Algorithm & 0.9981 & 3.9021  \\
\hline
Probability Distribution & $\mathcal{H}_1$-Algorithm & 0.5 & 3.8313  \\
\hline
Probability Distribution & $\mathcal{H}_2$-Algorithm & 0.52 & 4.14  \\
\hline
\end{tabular}
\caption{The minimum threshold required to achieve $90\%$ accuracy for each technique and algorithm and the number the species ruled out.}
\label{tab:90_acc}
\end{table}

\begin{table}[h!]
\centering
\begin{tabular}{|c|c|c|c|}
\hline
\textbf{Technique} & \textbf{Algorithm} & \textbf{Threshold} & \textbf{Number of Species Ruled Out}  \\
\hline
Class Prediction & $\mathcal{H}_1$-Algorithm & $-$ & $-$ \\
\hline
Class Prediction & $\mathcal{H}_2$-Algorithm & $-$ & $-$  \\
\hline
Probability Distribution & $\mathcal{H}_1$-Algorithm & 0.65 & 2.8038  \\
\hline
Probability Distribution & $\mathcal{H}_2$-Algorithm & 0.66 & 3.3  \\
\hline
\end{tabular}
\caption{The minimum threshold required to achieve $95\%$ accuracy for each technique and algorithm and the number the species ruled out.}
\label{tab:95_acc}
\end{table}

In Table \ref{tab:90_acc}, we see that the class prediction technique fails to reach a $90\%$ accuracy for any given threshold when using the $\mathcal{H}_1$-algorithm and requires a threshold of $0.9981$ to achieve such an accuracy. On the other hand, the probability distribution technique only requires a threshold of $0.5$ and $0.52$ for the $\mathcal{H}_1$ and $\mathcal{H}_2$-algorithms respectively. Moreover, we see that the $\mathcal{H}_2$-algorithm is able to rule out more species on average compared to the $\mathcal{H}_1$-algorithm. As such, we see that in the exchange between accuracy and the number of species ruled out that the $\mathcal{H}_2$-algorithm and probability distribution technique combination is best. We find similar results in Table \ref{tab:95_acc} when $95\%$ accuracy is desired, with the probability distribution technique and $\mathcal{H}_2$-algorithm outperforming the other combinations. 

\subsubsection{Feature Selection}

The results discussed in previous sections use all $16$ features to classify the unlabelled units. However, as with most real-world datasets, we have little knowledge as to which features may be irrelevant or redundant \cite{tang2014feature}. A feature is called irrelevant if it does not directly aide the classification problem, and a redundant feature does not provide new information. The removal of such features reduces the running time and has been shown to increase accuracy \cite{janecek2008relationship}. As such, there is a wide range of literature devoted to techniques on feature selection, see for instance \cite{chen2020selecting,dash1997feature,kwak2002input}. 

\begin{figure}[h!]
    \centering
    \includegraphics[width=0.5\textwidth]{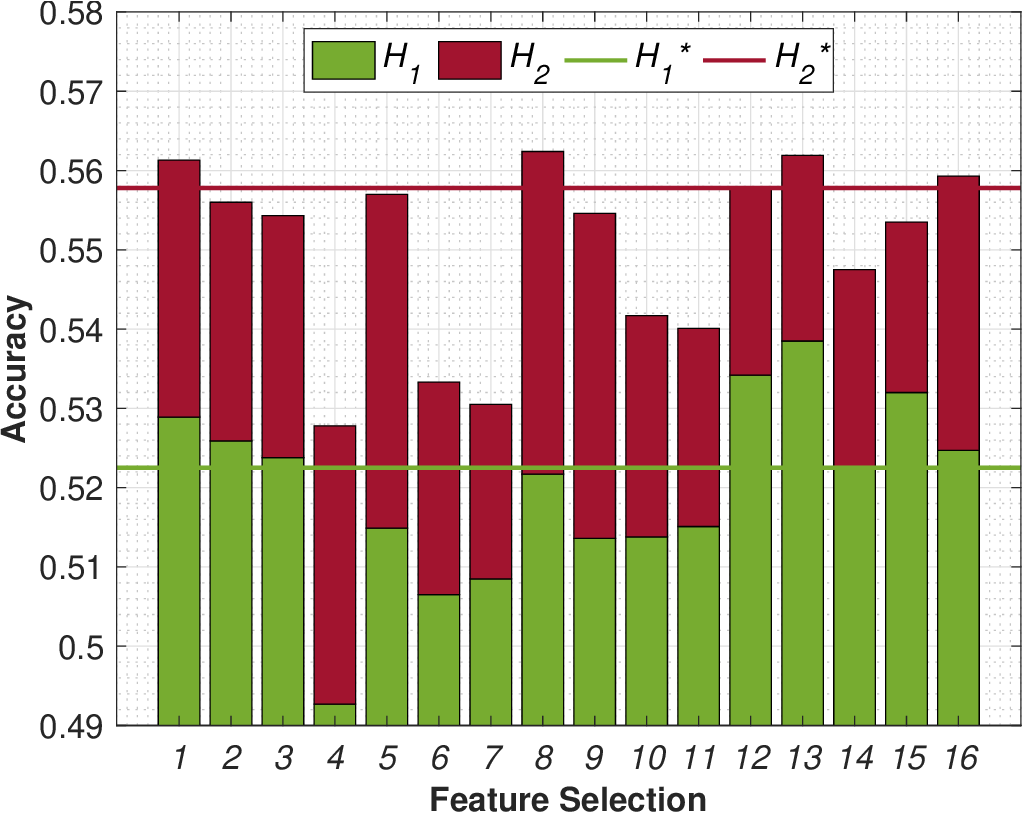}
    \caption{Adjusted and original accuracy scores for the $\mathcal{H}_1$ and $\mathcal{H}_2$-algorithms. Feature selection refers to which feature is removed, with the label coming from Table \ref{tab:grain_features}.}
    \label{fig:feat_selection_all}
\end{figure}

To explore the importance of feature selection, we consider the $\mathcal{H}_1$ and $\mathcal{H}_2$-algorithms when a particular feature is removed. When a feature is removed from the algorithm, we say the algorithm has been \textit{adjusted}. For example, if we remove \textit{Length} from the $\mathcal{H}_1$-algorithm, we refer to this as the \textit{Length}-adjusted $\mathcal{H}_1$-algorithm. The adjusted and original accuracy results are shown in Figure \ref{fig:feat_selection_all}. 

We first focus on the adjusted accuracy results for the $\mathcal{H}_1$-algorithm, drawing comparisons to the original accuracy and considering the most extreme changes. The most striking result seen in Figure \ref{fig:feat_selection_all} is the decrease in performance when \textit{Circularity} ($4$) is removed. This adjusted accuracy result demonstrates that \textit{Circularity} is a crucial feature for classification in this dataset. On the other hand, the removal of some features allowed the adjusted accuracy to increase compared to the original. The most prevalent of these features are \textit{Fourier Coefficients 6} and \textit{7} (12 and 13). This indicates that these two features, when removed, provide more clarity on classification. Therefore, we can view these features as being irrelevant as they do not aide in classification. 

The adjusted accuracy scores for the $\mathcal{H}_2$-algorithm also reveal some interesting results. When \textit{Circularity} ($4$) is removed we see a major decrease in accuracy compared to the original accuracy. As this also occurs in the $\mathcal{H}_1$-algorithm, we conclude that \textit{Circularity} aides in the classification of starch grains. Whereas \textit{Length} (1) and \textit{Fourier Coefficient 7} (13) when removed increase the accuracy. This behaviour is found in the $\mathcal{H}_1$-algorithm also, demonstrating that these features may be irrelevant or create confusion among the species identification. 

\begin{figure}[h!]
    \centering
    \includegraphics[width=0.5\textwidth]{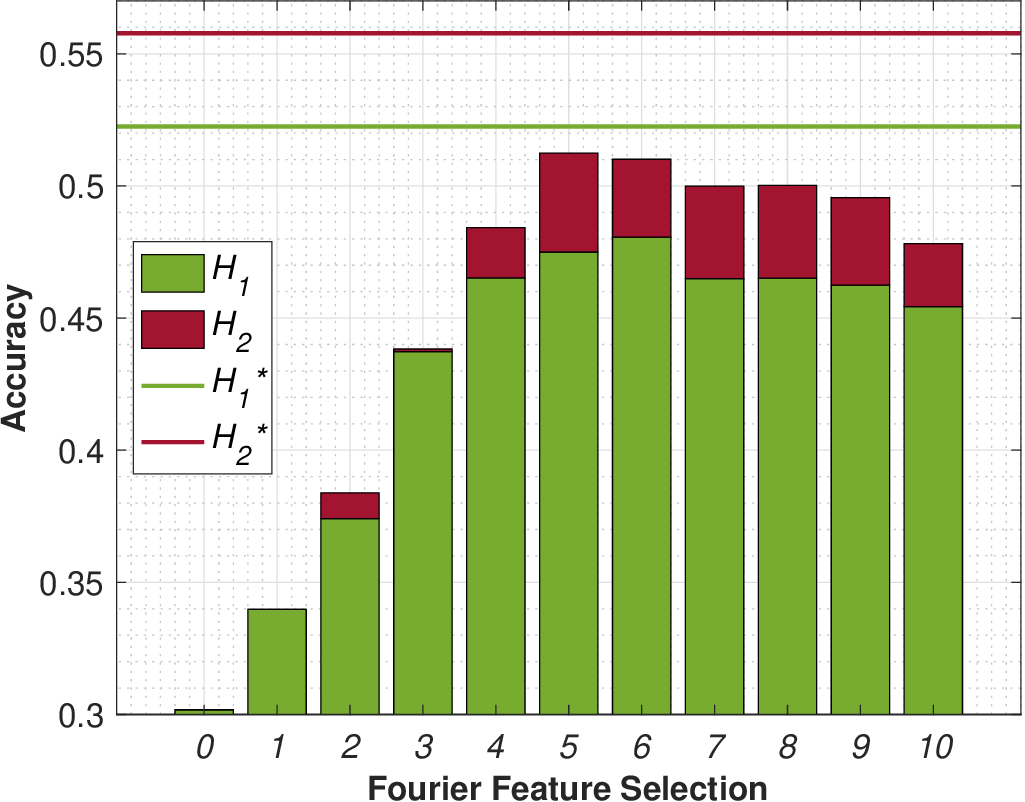}
    \caption{Adjusted and original accuracy scores for the $\mathcal{H}_1$ and $\mathcal{H}_2$-algorithms when only using Fourier Coefficients. The component(s) added into the algorithm are displayed on the $x$-axis.}
    \label{fig:Fourier_Comp_Acc}
\end{figure}

In \cite{coster2015starch}, the introduction of Fourier Coefficients enabled a rise in classification performance. Throughout this paper, we have included all Fourier Coefficients when implementing the classification algorithms. We now consider the benefits of not only including the Fourier Coefficients, but also how the higher order components aide in classification accuracy. Figure \ref{fig:Fourier_Comp_Acc} displays the accuracy of the $\mathcal{H}_1$ and $\mathcal{H}_2$-algorithms when only using the Fourier Coefficients and the accuracy adjustment when using more components. 

The results shown in Figure \ref{fig:Fourier_Comp_Acc} reveal some unexpected outcomes. We first notice that in all cases, we attain a decreased accuracy in comparison to the original classification where all features are included. This demonstrates that to achieve the optimal classification performance we must not ignore the features which are removed in this analysis (Features $1 - 5$ from Table \ref{tab:grain_features}). We then find that when including and adding Fourier Coefficients $0$ up to $6$, for both the $\mathcal{H}_1$ and $\mathcal{H}_2$-algorithms, there is an increase in accuracy. However, when including the remaining components, we see the accuracy stabilise and slightly decrease. From this, we conclude that the lower order components aide in classification performance while the higher order components do not.

\section{Conclusion}
In this work, we have introduced a novel hypergraph based multi-label classification algorithm known as the $\mathcal{H}_{\eta}$-algorithm. This algorithm makes use of the underlying multi-way interactions among features of datasets, which are often not directly studied in other hypergraph based classification algorithms. The algorithm is also constructed to produce robust results by considering a population of models, rather than a single, potentially biased and overfitting model. 

We exemplify the strong performance of the algorithm across two different datasets. The first being Fisher's Iris dataset, which is a well-studied benchmark classification dataset and the other being a complex starch grain dataset. In both cases we achieve promising results, with the $\mathcal{H}_{\eta}$-algorithm performing similarly to the random forest on Fisher's Iris dataset. While the $\mathcal{H}_{\eta}$-algorithm out-performed the random forest on the complex starch grain dataset for ${\eta} =2$ and ${\eta} = 3$.

We were also able to show that the introduction of a threshold during the classification process increased performance. In particular, we were able to use the probability distributions encoded by the hyperedges to rule out species while attaining high accuracy results ($\geqslant 0.9$).

Overall, the hypergraph foundations used in the $\mathcal{H}_{\eta}$-algorithm highlight salient multi-way interactions within datasets. While the population of hypergraphs give a robust framework for classification of unseen data points. These two concepts combine to give a novel and powerful classification algorithm for both simple and complex datasets. 

\newpage
\section*{Appendix A}
\subsubsection{Algorithm}
\label{subsub:alg}

In this section, we provide an algorithm perspective for predicting the class of an unlabelled unit based on the above model. We refer to this as Algorithm \ref{subsub:alg}. Algorithm \ref{subsub:alg} considers the unlabelled unit $v$, and a given training set of units. We note here that \textbf{Output ($\star$)} introduces a decision threshold condition which is discussed in Section \ref{subsub:threshold_classification_performance}.

\begin{align*}
& \textrm{ \textbf{Input}} && \text{Let $v'$ be the raw data of the unlabelled unit $v$,} \\[0ex]
&  && v' = (v'_1,\dots,v'_m).  \\[1ex]
& \textrm{ \textbf{1. Normalise}} && \text{Normalise $v'$ ($\S$ \ref{sec:norm_disc}),} \\[0ex]
& && \hat{v} = \Bigg(\frac{v_1' - \mu_1}{\sigma_1},\dots,\frac{v_m' - \mu_m}{\sigma_m} \Bigg) = (\hat{v}_1,\dots,\hat{v}_m). \\[1ex]
& \textrm{ \textbf{2. Parameter Selection}} && \text{Choose pair $\ell \sim U(0.2,1.5)$ and $\alpha \sim U(-\frac{1}{2},\frac{1}{2})$.} \\[1ex]
& \textrm{ \textbf{3. Discretise}} && \text{Discretise $\hat{v}$ ($\S$ \ref{sec:norm_disc}),} \\[0ex] 
& && v = \Bigg( \Bigg\lceil \frac{\hat{v}_1 - \alpha}{\ell} \Bigg\rceil,\dots,\Bigg\lceil \frac{\hat{v}_m - \alpha}{\ell} \Bigg\rceil \Bigg) = (v_1,\dots,v_m). \\[1ex] 
& \textrm{ \textbf{4. Incidence Tuple}} && \text{Transform $v$ to an incidence tuple using \eqref{incidence_tuple} ($\S$ \ref{subsub:predict}),} \\[0ex] 
&  && \nu = (\nu_1,\dots,\nu_{\tau m}). \\[1ex] 
& \textrm{ \textbf{5. Distribution Array}} && \text{Calculate the distribution array based on $\nu$ ($\S$ \ref{subsub:predict}),} \\[0ex] 
& && V = \diag(\nu) \cdot W(P_{\ell,\alpha}). \\[1ex] 
& \textrm{ \textbf{6. Predict}} && \text{Predict class via $\overline{\bm{V}}$, the column means of $V$ ($\S$ \ref{subsub:predict});} \\[0ex]
&  && f(V) = \argmax\Big(\overline{\bm{V}} \Big) \text{ gives the class prediction.} \\[1ex]
& \textrm{ \textbf{7. Population of Models}} && \text{Repeat \textbf{2} - \textbf{6} with variations of } \ell \text{ and } \alpha.  \\[1ex]
& \textrm{ \textbf{8. Population of Predictions}} && \text{Store population of predictions in } p_v \text{ ($\S$ \ref{subsub:pop})} \\[1ex]
& \textrm{ \textbf{9. Final Prediction}} && \text{Calculate the most frequent class in $p_v$ ($\S$ \ref{subsub:pop}),} \\[0ex]
&  && g(p_{v}) = \argmax_k P(p_{v} = k). \\[1ex]
& \textrm{ \textbf{Output}} && v \text{ is predicted to be class } g(p_v). \\
& \textrm{ \textbf{Output ($\star$)}} && v \text{ is predicted to be class } g(p_v) \iff P(p_v = k) > 0.75.
\end{align*}

\newpage
\section*{Appendix B}
\begin{table}[h!]
    \centering
    \resizebox{0.5\textwidth}{!}{\begin{tabular}{|c|c|c|c|c|c|}
    \hline
    Species & Fold 1 & Fold 2 & Fold 3 & Fold 4 & Fold 5 \\
    \hline
    $DK$ & 25 & 26 & 33 & 18 & 21 \\
    \hline
    $AA$ & 22 & 17 & 26 & 21 & 19 \\
    \hline
    $AV$ & 25 & 19 & 27 & 30 & 20 \\
    \hline
    $BM$ & 25 & 27 & 28 & 30 & 24 \\
    \hline
    $EE$ & 26 & 28 & 30 & 35 & 35 \\
    \hline
    $YA$ & 28 & 28 & 29 & 23 & 34 \\
    \hline
    $BP$ & 31 & 33 & 32 & 30 & 39 \\
    \hline
    \end{tabular}}
    \caption{The number of units per species used in each partition for the $5$-fold cross validation.}
    \label{tab:5fold_values}
\end{table}

\begin{table}[h!]
\centering
\resizebox{0.5\textwidth}{!}{\begin{tabular}{ c|c|c|c|c|c|c|c|c| } 
\multicolumn{2}{c}{} & \multicolumn{7}{c}{\textbf{True Class}} \\
\cline{3-9}
\multicolumn{2}{c|}{} & $DK$ & $AA$ & $AV$ & $BM$ & $EE$ & $YA$ & $BP$ \\
\cline{2-9}
\multirow{7}{*}{\textbf{Output}} & $DK$ & 102 & 11 & 3 & 1 & 5 & 1 & 0 \\
\cline{2-9}
\multirow{7}{*}{\textbf{Class}} & $AA$ & 1 &	86	& 14 & 	1 & 0 &	1 &	2 \\
\cline{2-9}
& $AV$ & 0 & 38 & 47 & 9 & 1 & 1 & 25 \\
\cline{2-9}
& $BM$ & 1 &	0 & 17 & 81 & 11 & 16 &	8 \\
\cline{2-9}
& $EE$ & 38 & 8 & 6 & 47 & 38 & 17 &	0 \\
\cline{2-9}
& $YA$ & 13 & 6 & 6 & 32 & 16 & 69 &	0 \\
\cline{2-9}
& $BP$ & 8 &	30 & 32 & 16 & 6 & 3 & 70 \\
\cline{2-9}
\end{tabular}}
\caption{Confusion matrix for $\mathcal{H}_1$-algorithm.}
\label{tab:Conf_H1}
\end{table}

\begin{table}[h!]
\centering
\resizebox{0.5\textwidth}{!}{\begin{tabular}{ c|c|c|c|c|c|c|c|c| } 
\multicolumn{2}{c}{} & \multicolumn{7}{c}{\textbf{True Class}} \\
\cline{3-9}
\multicolumn{2}{c|}{} & $DK$ & $AA$ & $AV$ & $BM$ & $EE$ & $YA$ & $BP$ \\
\cline{2-9}
\multirow{7}{*}{\textbf{Output}} & $DK$ & 109 & 8 & 2 & 0 & 3 & 1 & 0 \\
\cline{2-9}
\multirow{7}{*}{\textbf{Class}} & $AA$ & 1 &	85 & 15 & 1 & 0 & 1 & 2 \\
\cline{2-9}
& $AV$ & 0 & 32 & 53 & 9 & 1 & 0 & 26 \\
\cline{2-9}
& $BM$ & 2 &	0 & 14 & 81 & 13 & 17 &	7 \\
\cline{2-9}
& $EE$ & 42 & 6 & 5 & 37 & 46 & 17 &	1 \\
\cline{2-9}
& $YA$ & 14 & 4 & 5 & 32 & 14 & 72 &	1 \\
\cline{2-9}
& $BP$ & 10 & 27 & 29 & 12 & 4 & 3 & 80 \\
\cline{2-9}
\end{tabular}}
\caption{Confusion matrix for $\mathcal{H}_2$-algorithm.}
\label{tab:Conf_H2}
\end{table}

\begin{table}[h!]
\centering
\resizebox{0.5\textwidth}{!}{\begin{tabular}{ c|c|c|c|c|c|c|c|c| } 
\multicolumn{2}{c}{} & \multicolumn{7}{c}{\textbf{True Class}} \\
\cline{3-9}
\multicolumn{2}{c|}{} & $DK$ & $AA$ & $AV$ & $BM$ & $EE$ & $YA$ & $BP$ \\
\cline{2-9}
\multirow{7}{*}{\textbf{Output}} & $DK$ & 114 & 4 & 2 & 0 & 2 & 1 & 0 \\
\cline{2-9}
\multirow{7}{*}{\textbf{Class}} & $AA$ & 1 &	83	& 16 & 1 & 0 & 1 & 3 \\
\cline{2-9}
& $AV$ & 1 & 30 & 56 & 6 & 1 & 0 & 27 \\
\cline{2-9}
& $BM$ & 2 &	0 & 15 & 79 & 14 & 17 &	7 \\
\cline{2-9}
& $EE$ & 45 & 6 & 6 & 33 & 44 & 19 &	1 \\
\cline{2-9}
& $YA$ & 14 & 3 & 4 & 34 & 19 & 67 &	1 \\
\cline{2-9}
& $BP$ & 10 & 22 & 30 & 11 & 4 & 3 & 85 \\
\cline{2-9}
\end{tabular}}
\caption{Confusion matrix for $\mathcal{H}_3$-algorithm.}
\label{tab:Conf_H3}
\end{table}

\begin{table}[h!]
\centering
\resizebox{0.5\textwidth}{!}{\begin{tabular}{ c|c|c|c|c|c|c|c|c| } 
\multicolumn{2}{c}{} & \multicolumn{7}{c}{\textbf{True Class}} \\
\cline{3-9}
\multicolumn{2}{c|}{} & $DK$ & $AA$ & $AV$ & $BM$ & $EE$ & $YA$ & $BP$ \\
\cline{2-9}
\multirow{7}{*}{\textbf{Output}} & $DK$ & 104 & 5 & 0 & 0 & 9 & 6 & 0 \\
\cline{2-9}
\multirow{7}{*}{\textbf{Class}} & $AA$ & 2 &	55	& 27 & 3 & 0 & 2 & 11 \\
\cline{2-9}
& $AV$ & 1 & 17 & 43 & 6 & 2 & 4 & 42 \\
\cline{2-9}
& $BM$ & 1 &	2 & 6 & 70 & 26 & 22 & 9 \\
\cline{2-9}
& $EE$ & 35 & 9 & 0 & 31 & 66 & 14 &	1 \\
\cline{2-9}
& $YA$ & 8 & 1 & 2 & 24 & 26 & 79 & 2 \\
\cline{2-9}
& $BP$ & 5 & 16 & 30 & 14 & 3 & 5 & 94 \\
\cline{2-9}
\end{tabular}}
\caption{Confusion matrix for random forest algorithm.}
\label{tab:Conf_RF}
\end{table}

\begin{table}[h!]
\centering
\resizebox{0.5\textwidth}{!}{\begin{tabular}{ c|c|c|c|c|c|c|c|c| } 
\multicolumn{2}{c}{} & \multicolumn{7}{c}{\textbf{True Class}} \\
\cline{3-9}
\multicolumn{2}{c|}{} & $DK$ & $AA$ & $AV$ & $BM$ & $EE$ & $YA$ & $BP$ \\
\cline{2-9}
\multirow{7}{*}{\textbf{Output}} & $DK$ & 96 & 4 & 1 & 0 & 0 & 0 & 0 \\
\cline{2-9}
\multirow{7}{*}{\textbf{Class}} & $AA$ & 1 &	61	& 4 & 0 & 0 & 0 & 0 \\
\cline{2-9}
& $AV$ & 0 & 14 & 26 & 2 & 0 & 0 & 16 \\
\cline{2-9}
& $BM$ & 1 &	0 & 3 & 46 & 3 & 3 & 4 \\
\cline{2-9}
& $EE$ & 33 & 3 & 1 & 14 & 15 & 1 &	0 \\
\cline{2-9}
& $YA$ & 7 & 2 & 0 & 13 & 3 & 37 & 0 \\
\cline{2-9}
& $BP$ & 6 & 14 & 16 & 6 & 0 & 0 & 59 \\
\cline{2-9}
\end{tabular}}
\caption{Confusion matrix for $\mathcal{H}_3$-algorithm with threshold of $0.75$.}
\label{tab:Conf_H3_threshold}
\end{table}

\newpage
\bibliographystyle{abbrv}
\bibliography{references}

\end{document}